\documentclass[10pt,twocolumn,letterpaper]{article}
\usepackage{wacv}
\usepackage[square,comma,sort,numbers]{natbib}
\usepackage{graphicx}
\usepackage{amsmath}
\usepackage{amssymb}
\usepackage{booktabs}
\usepackage{multirow}
\usepackage{stmaryrd}
\usepackage{algorithm}
\usepackage{algpseudocode}
\usepackage{pifont}
\usepackage{color, colortbl}
\usepackage{enumitem,kantlipsum}
\usepackage[dvipsnames]{xcolor}
\usepackage[pagebackref,breaklinks,colorlinks]{hyperref}
\newcommand{\cmark}{\ding{51}}
\newcommand{\xmark}{\ding{55}}
\definecolor{Gray}{gray}{0.95}

\usepackage[capitalize]{cleveref}
\crefname{section}{Sec.}{Secs.}
\Crefname{section}{Section}{Sections}
\Crefname{table}{Table}{Tables}
\crefname{table}{Tab.}{Tabs.}

\DeclareMathOperator*{\argmin}{arg\,min}
\definecolor{coolblack}{rgb}{0.0, 0.23, 0.64}
\newcommand{\ak}[1]{\textcolor{red}{#1}}
\newcommand{\jnkc}[1]{\textcolor{coolblack}{#1}}

\usepackage{pict2e,picture,graphicx}
\makeatletter
\DeclareRobustCommand{\Arrow}[1][]{
\check@mathfonts
\if\relax\detokenize{#1}\relax
\settowidth{\dimen@}{$\m@th\rightarrow$}
\else
\setlength{\dimen@}{#1}
\fi
\sbox\z@{\usefont{U}{lasy}{m}{n}\symbol{41}}
\begin{picture}(\dimen@,\ht\z@)
\roundcap
\put(\dimexpr\dimen@-.7\wd\z@,0){\usebox\z@}
\put(0,\fontdimen22\textfont2){\line(1,0){\dimen@}}
\end{picture}
}
\makeatother

\newcommand{\expectation}{\mathop{\mathbb{E}}}

\newcommand{\PreserveBackslash}[1]{\let\temp=\\#1\let\\=\temp}
\newcommand{\improvement}[1]{\small{\textbf{\color{ForestGreen}{(+#1)}}}}
\newcolumntype{C}[1]{>{\PreserveBackslash\centering}p{#1}}
\newcolumntype{R}[1]{>{\PreserveBackslash\raggedleft}p{#1}}
\newcolumntype{L}[1]{>{\PreserveBackslash\raggedright}p{#1}}

\def\wacvPaperID{443}
\def\confName{WACV}
\def\confYear{2024}

\begin{document}
\title{Aligning Non-Causal Factors for \\ Transformer-Based Source-Free Domain Adaptation}

\author{
Sunandini Sanyal\thanks{Equal Contribution} \quad
Ashish Ramayee Asokan\footnotemark[1] \quad Suvaansh Bhambri \quad Pradyumna YM \\ Akshay Kulkarni \quad Jogendra Nath Kundu \quad R Venkatesh Babu \\ 
Vision and AI Lab, Indian Insitute of Science, Bengaluru \\
}
\maketitle

\begin{abstract}
   Conventional domain adaptation algorithms aim to achieve better generalization by aligning only the task-discriminative causal factors between a source and target domain. However, we find that retaining the spurious correlation between causal and non-causal factors plays a vital role in bridging the domain gap and improving target adaptation. Therefore, we propose to build a framework that disentangles and supports causal factor alignment by aligning the non-causal factors first. We also investigate and find that the strong shape bias of vision transformers, coupled with its multi-head attention, make it a suitable architecture for realizing our proposed disentanglement. Hence, we propose to build a Causality-enforcing Source-Free Transformer framework (C-SFTrans\footnote{Project Page: \url{https://val.cds.iisc.ac.in/C-SFTrans/}}) to achieve disentanglement via a novel two-stage alignment approach: a) non-causal factor alignment: non-causal factors are aligned using a style classification task which leads to an overall global alignment, b) task-discriminative causal factor alignment: causal factors are aligned via target adaptation. We are the first to investigate the role of vision transformers (ViTs) in a privacy-preserving source-free setting. Our approach achieves state-of-the-art results in several DA benchmarks. 
\end{abstract}

\section{Introduction}
Machine learning models often fail to generalize well in scenarios where the test data distribution (\textit{source domain}) differs a lot from the training data distribution (\textit{target domain}). In practice, a model often encounters data from unseen domains \ie \textit{domain shift}. This leads to a poor deployment performance, which critically impacts many real-world applications such as autonomous driving \cite{cordts2016cityscapes}, surveillance systems \cite{723792}, etc. 

\begin{figure}
\centering
        \includegraphics[width=1.0\linewidth]{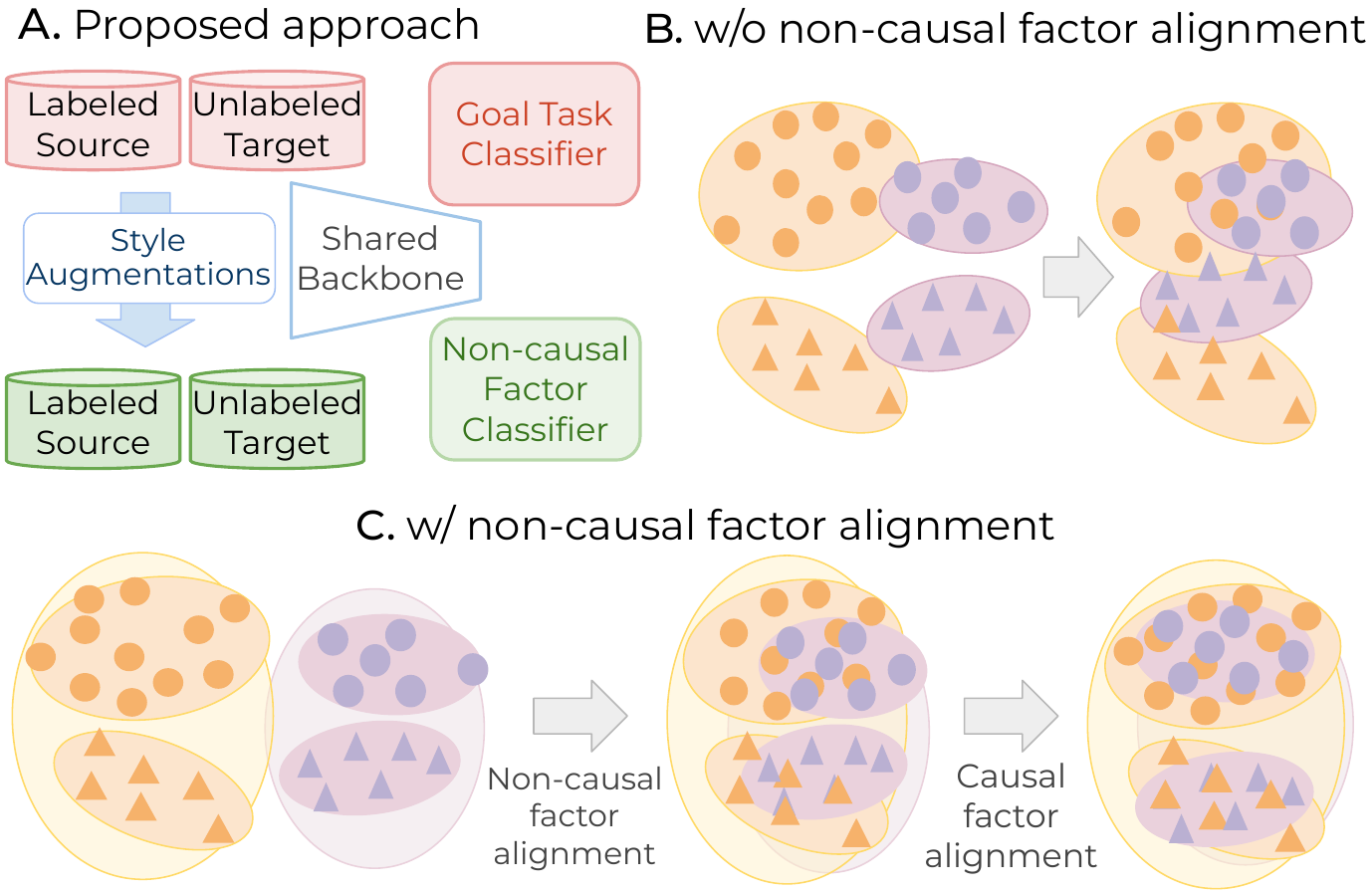}
        \caption{\textbf{A. Proposed method.} We incorporate non-causal factors to learn domain-invariant representations using a subsidiary non-causal factor classification task.  \textbf{B. w/o Non-Causal Factor Alignment.} Conventional domain-invariance methods aim to align only causal factors, leading to sub-optimal alignment between the source and target domain. \textbf{C. w/ Non-Causal Factor Alignment.} Non-causal factor alignment improves global alignment, leading to better task-discriminative causal factor alignment.}
        \vspace{-1em}
       \label{fig:teaser}
\end{figure}

Unsupervised domain adaptation (UDA) methods \cite{ganin2015unsupervised} aim to address the challenges of domain shift by learning the task knowledge of the labeled source domain and adapting to an unlabeled target domain. However, these works \cite{ganin2016domain} require joint access to the source and target data. Such a constraint is highly impractical as data sharing is usually restricted in most real-life applications due to privacy concerns. Hence, in this work, we focus on the practical problem setting of source-free domain-adaptation (SFDA) \cite{kundu2020universal} where a \textit{vendor} trains a source model and shares only the source model with a \textit{client} for target adaptation.

Conventional DA works \cite{ganin2016domain} aim to learn domain-invariant representations by aligning only the task-related features between the source and target domain. We refer to these features as \textit{causal factors} that heavily influence the goal task. We also denote factors that capture contextual information as \textit{non-causal factors}. Causal factor alignment leads to a low target error, thereby improving the adaptation performance, as shown theoretically by \citet{ben2010theory}. But these methods require concurrent access to the source and target domain data, which is impractical in restricted data sharing scenarios of SFDA. Further, causal and non-causal factors are spuriously correlated \cite{sun2021recovering} and this correlation may break when a domain shift occurs. Hence, in our work, we propose to retain this spurious correlation through disentanglement and learning of both causal and non-causal factors.

Motivated by the remarkable success of vision transformer architectures \cite{khan2021transformers}, we propose to explore the possibility of disentanglement and alignment of causal/non-causal factors using transformers. Recent domain-invariance-based SFDA works \cite{wang2022exploring, kundu2022balancing} have been found to be highly effective on convolution-based architectures. However, in our analysis, we find that a simple vision transformer (ViT) baseline outperforms the state-of-the-art CNN-based methods, implying that transformers are highly robust to domain shifts \cite{naseer2021intriguing}. Secondly, we observe that the domain-invariance methods do not significantly impact the performance of vision transformers due to their inherent shape bias \cite{park2022vision}. Based on these observations, we propose to leverage ViTs for a realizable disentanglement of causal and non-causal factors. Hence, in our work, we seek an answer to an important question, \textit{``How do we develop a framework for disentanglement using vision transformers to retain the spurious correlation between causal and non-causal factors, in the challenging source-free setting?"}. 

Conventional approaches \cite{ganin2016domain, wang2022exploring, lv2022causality} completely ignore the non-causal factors and align only the causal factors, which leads to sub-optimal alignment of the class-specific clusters as shown in Fig.\ \ref{fig:teaser}\ak{B}. This results in poor performance, especially in cases of large domain gap between the source and target. Hence, we propose to first align the non-causal factors, which leads to an overall global alignment between the source and target domain. We then align the goal task-discriminative causal factors that optimally aligns the local class-specific clusters (Fig.\ \ref{fig:teaser}\ak{C}). Hence, in our work, we pinpoint that \textit{``aligning the non-causal factors is crucial for improving target adaptation performance"}.
 
We next seek to devise a framework that explicitly guides the process of disentanglement and alignment. We develop a novel framework - \textit{Causality-enforcing  Source-Free Transformers} (\textbf{C-SFTrans}) that comprises of two stages, \textbf{a)} Non-causal factor alignment, and \textbf{b)} Task-discriminative causal factor alignment. To enable non-causal factor alignment, we propose a subsidiary task of non-causal factor classification for a source-free setting. Prior works \cite{chen2022principle} show that multi-head self-attentions in transformers focus on redundant features. To facilitate the separate learning of diverse non-causal factors and causal factors, we utilize this inherent potential of transformers to designate non-causal and causal attention heads for training. We propose a novel \textit{Causal Influence Score Criterion} to perform the head selection. The two stages of non-causal factor alignment and task-discriminative causal factor alignment are performed alternatively to achieve domain-invariance. We outline the major contributions of our work:

\begin{itemize}
    \item To the best of our knowledge, we are the first to explore vision transformers (ViTs) for a practical source-free DA setting. We investigate domain-invariance in vision transformers and provide insights to improve target adaptation via non-causal factor alignment. 
    \item We propose a novel two-stage disentanglement and alignment framework \textit{Causality-enforcing  Source-Free Transformers} (\textbf{C-SFTrans}) to preserve the spurious correlation between causal and non-causal factors.
    
    \item We define a novel attention head selection criterion, \textit{Causal Influence Score Criterion}, to select attention heads for non-causal factor alignment. We also introduce a novel Style Characterizing Input (SCI) to further aid the head selection.

    \item We achieve state-of-the-art results on several source-free benchmarks of single-source and multi-target DA.
    
\end{itemize}

\begin{figure*}
\centering
        \includegraphics[width=1.0\linewidth]{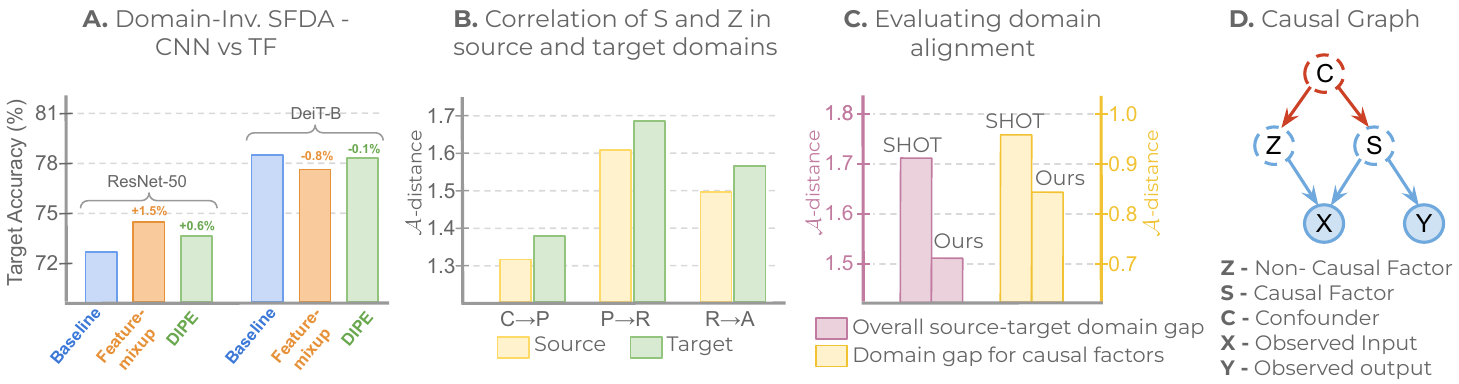}
        \caption{\textbf{A.}
        SOTA domain-invariance-based DA works, Feature-Mixup \cite{kundu2022balancing} and DIPE \cite{wang2022exploring}, do not improve over the baseline for vision transformers despite large gains for CNNs. \textbf{B.} We observe that correlation $S$ and $Z$ in preserved after target adaptation. (Office-Home) \textbf{C.} With our proposed style task training, the overall source-target domain gap (pink) is lower indicating better domain-invariance. Further, we observe a lower domain gap considering only causal factors (yellow) indicating that non-causal alignment helps causal alignment. \textbf{D.} Causal Graph representing causal factors $S$ and non-causal factors $Z$, which are spuriously correlated via confounder $C$. }
        \vspace{-1em}
       \label{fig:pof}
\end{figure*}

\section{Related works}
\label{sec:rel_works}

\noindent
\textbf{Unsupervised Domain Adaptation.} Unsupervised Domain Adaptation (DA) aims to adapt a source-trained model to a given target domain. DA approaches can be broadly classified into two categories: 1) methods using generative models \cite{russo2018source, sankaranarayanan2018generate, liu2016coupled} to create synthetic target-like images for adaptation. 2) methods focusing on aligning the source and target feature distributions using statistical distance measures on the source/target features \cite{NRC, Peng2019DomainAL, saito2018maximum, saenko2010adapting}, and adversarial training \cite{hu2018duplex, hoffman2018cycada, long2017deepJAN}. Recent works \cite{roy2022uncertainty, kundu2022concurrent, qu2022bmd, lee2022confidence} address a more restrictive and privacy-preserving setting of \emph{Source-Free Domain Adaptation} (SFDA), where the source data is inaccessible during target adaptation. SFDA works of SHOT \cite{SHOT} and SHOT++ \cite{SHOT++} use pseudo-labeling and information maximization to align the source and target domains. 
Our work also addresses the challenging SFDA setting, intending to improve the adaptation performance. 

\noindent
\textbf{Transformers for Domain Adaptation.} Despite the success of Vision Transformers (ViTs) \cite{dosovitskiy2020image} in several vision tasks, their application to DA has been relatively less explored. TransDA \cite{transDA} improves the model generalizability by incorporating the transformer's attention module in a convolutional network. CDTrans \cite{xu2021cdtrans} proposes a transformer framework comprising three weight-sharing branches for cross-attention and self-attention using the source and target samples, while SSRT \cite{sun2022safe} introduces a self-training strategy that uses perturbed versions of the target samples to refine the ViT backbone. TVT \cite{yang2021tvt} improves the transferability of ViTs through adversarial training.

\noindent
\textbf{Domain Invariance for Domain Adaptation.} These methods aim to learn domain-invariant feature representations between the source and target domains. SHOT \cite{SHOT} and SHOT++ \cite{SHOT++} prevent updates to the source hypothesis, which enables the feature extractor to learn domain-invariant representations. Feature-Mixup \cite{kundu2022balancing} constructs an intermediate domain whose representations preserve the task discriminative information while being domain-invariant. In contrast, DIPE \cite{wang2022exploring} trains the domain-invariant parameters of the source-trained model rather than learning domain-invariant representations between the domains.

\noindent
\textbf{Causal representation learning.} Causality mechanisms \cite{scholkopf2021toward} focus on learning invariant representations and recovering causal features \cite{wang2022contrastive, lv2022causality} that improve the model's generalizability. Some works \cite{ye2022alleviating, kim2020learning, chen2021unsupervised} attempt this via texture-invariant representation learning. However, such works are less effective as they align only the causal factors towards improving generalization. In contrast, we propose a novel way of learning domain-invariant representations by taking into account both the non-causal and causal factors in the target domain to achieve the best adaptation performance.

\section{Approach}
\label{sec:approach}

\subsection{Preliminaries}

\noindent
\textbf{Problem Setting.}
We consider the problem setting of closed-set DA, with a labeled source domain dataset ${{{\mathcal{D}}_s = \{( x_s, y_s) : x_s \!\in\! {\mathcal{X}}, y_s \!\in\! {\mathcal{C}}_g\}}}$ where ${\mathcal{X}}$ is the input space and $\mathcal{C}_g$ is the class label set. The unlabeled target dataset is denoted as ${{\mathcal{D}}_t = \{ x_t : x_t \!\in \!{\mathcal{X}}\}}$. The task of DA is to predict the label for each target sample $x_t$ from the label space $\mathcal{C}_g$. Following \citet{xu2021cdtrans}, we use a vision transformer backbone ViT-B \cite{dosovitskiy2020image} as the feature extractor, denoted as $f:\mathcal{X}\!\to\! \{\mathcal{Z}_c, \mathcal{Z}_1, \mathcal{Z}_2, ... \mathcal{Z}_{N_P}\}$. $\mathcal{Z}_c$ denotes the class token feature-space and $\mathcal{Z}_1, \mathcal{Z}_2, ...\mathcal{Z}_{N_P}$ are patch token feature-spaces. $N_P$ denotes the number of patches. We train a goal task classifier $f_g$ on the class-token as $f_g: \mathcal{Z}_c\to \mathcal{C}_g$. In this work, we operate under the vendor-client paradigm of source-free domain adaptation \cite{kundu2020universal} where a vendor trains a source model on the labeled source domain dataset and shares the model with the client. The client, on the other hand, trains the model with the unlabeled target domain data for target adaptation.

\noindent
\textbf{Causal Model.}
Let $X$ denote the input variable and $Y$ denote the output variable or label. We represent the structural causal graph $G$ as shown in Fig.\ \ref{fig:pof}\ak{D} where we introduce latent variables $S$ and $Z$ to capture the generative concepts (\eg object shape, backgrounds, textures) that lead to the observed variables $X$ and $Y$. We explicitly separate causal factors $S$ that causally influence the class-label $Y$ (\eg shape) and non-causal factors $Z$ which refers to contextual information (like background, texture, \etc). We also assume that $S$ is spuriously correlated with $Z$ shown through an unobserved confounder $C$. This spurious correlation may vary across domains. For instance, in a source-free DA setting, the source domain might have a ``ball" in a ``football field", while the target domain may have a ``ball" in ``table tennis". For simplicity, we use Fig.\ 2(b) of Sun \etal \cite{sun2021recovering} where domain shifts are represented as changes in the probability distributions $P(C)$ or $P(S, Z | C)$.

\noindent
\textbf{Analyzing Causality in Source-Free DA.}
Since $S$ and $Z$ are correlated spuriously, a source model trained on source domain data $X$ inherits this spurious correlation in the form of bias (\eg contextual information of objects and scenes occurring together in the source domain). In the target domain, this spurious correlation changes, leading to worsened performance of the source model. In our work, we aim to retain the spurious correlation between $S$ and $Z$ in the source-free DA setting when domain shift occurs. We construct an explicitly disentangled network to model the causal factors $S$ and the non-causal factors $Z$ through separate learning objectives. When the domain changes, we leverage the disentanglement and our designed training objectives to retain the $P(S, Z|C)$ correlation.

\vspace{1mm}
\noindent
We begin by first investigating the impact of existing CNN-based DA methods on vision transformers (ViTs) and draw insights 
towards achieving the desired disentanglement.

\subsection{Exploring domain-invariance for ViTs}
Motivated by the impressive performance of vision transformers (ViT) across several vision tasks \cite{khan2021transformers}, we propose to investigate the impact of domain-invariance methods on vision transformers in the highly practical and privacy-oriented source-free setting. A few recent works \cite{wang2022exploring, kundu2022balancing} propose to learn domain-invariant representations without access to source data. However, we find that these approaches mainly work only with CNN architectures. We first extend two state-of-the-art domain-invariance works, Feature Mixup \cite{kundu2022balancing} and DIPE \cite{wang2022exploring}, for transformers (see Fig.\ \ref{fig:pof}\ak{A}). We observe that these works exhibit marginal drops compared to a baseline SHOT \cite{SHOT} for transformers, although they have significant gains for CNNs.

But why does this happen? Recent works \cite{raghu2021vision} 
show that ViTs incorporate more global information than CNNs. This is because the multi-head self-attention captures causal factors via shape bias \cite{park2022vision}, while convolutions capture non-causal factors via texture bias. As convolutions inherently capture texture bias, domain-invariance methods perform well as they align causal factors and improve the shape bias in CNNs \cite{kundu2022balancing}. Since transformers implicitly possess a stronger shape bias, it is more robust to domain shifts \cite{naseer2021intriguing} and existing methods do not improve the adaptation performance significantly, unlike in CNNs. Based on these observations, we hypothesize that ViTs can better model the causal factors and can thus accommodate the disentanglement of non-causal factors, which are easier to learn. However, the question remains, \textit{``How can we leverage such a disentanglement in ViTs to retain the spurious correlation between causal and non-causal factors, while operating under the challenging yet practical source-free DA constraints?" }

Next, we propose a method for disentanglement in ViTs for modeling the causal and non-causal factors.

\begin{figure*}
\centering
        \includegraphics[width=1.0\linewidth]
        {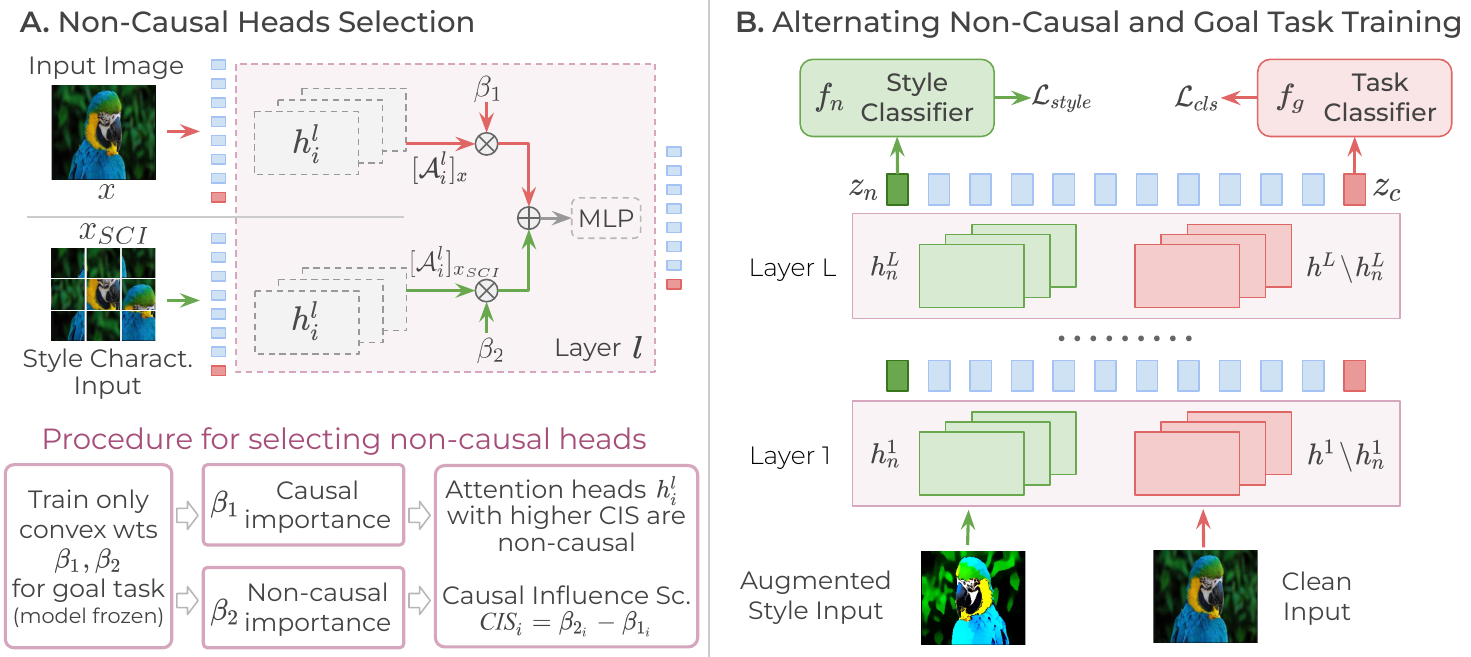}
        \caption{
        \textbf{A.}
        We estimate the causal importance of each attention head in a block by training convex weights $\beta_1$ and $\beta_2$ that combine the output of the attention heads for normal input $x$ and style characterizing input $x_{SCI}$. A higher $\beta_2$ indicates that the attention head gave more importance to non-causal style factors and can be chosen as a non-causal head.
        \textbf{B.}
        The selected non-causal heads $h_n$ are updated to train the style classifier $f_n$ with the style classification loss $\mathcal{L}_{style}$ via a style token $z_n$. The task classifier is trained by updating only causal heads $\mathcal{H}\!\setminus\!h_n$ with the task classification loss $\mathcal{L}_{cls}$ via the class token $z_c$. The two steps are executed alternately.}
        \vspace{-1em}
       \label{fig:approach}
\end{figure*}

\subsection{Non-causal factor alignment for source-free DA}
Conventional approaches ignore non-causal factors while aligning only the causal factors. We know that the spurious correlation between causal and non-causal factors can heavily influence the classification performance \cite{sun2021recovering}, especially in scenarios with a large source-target domain gap. Hence, disentangling and aligning non-causal factors can help bridge the domain gap to a large extent, especially for ViT architectures with a strong shape bias. Therefore, we come up with the following insight of effectively utilizing the non-causal factors to enable better alignment of causal factors in the challenging source-free DA setting.

\vspace{1mm}
\noindent
\textbf{Insight 1.\ (Non-causal factor alignment positively influences causal factor alignment)} 
\textit{Aligning non-causal factors leads to an overall global alignment between the source and target domain for source-free DA settings. In other words, it implicitly improves the alignment of causal factors as well, thereby improving the adaptation performance.
}

\noindent
\textbf{Remarks.}
Non-causal factor alignment forces the model to focus on the non-causal factors. Causal factors modeled through the inherent shape bias of transformers lead to substantial gains over CNNs (as shown in Fig.\ \ref{fig:pof}\ak{A}). However, aligning the residual non-causal factors can further improve the overall global alignment which helps to preserve the correlation between causal and non-causal factors in the target domain.  We demonstrate this phenomenon in Fig.\ \ref{fig:pof}\ak{C} (pink bars), where our non-causal factor alignment improves the overall global alignment between the two domains, leading to a significant reduction in the domain gap between the source and target domain. This shows that non-causal factor alignment positively influences causal factor alignment as the spurious correlation between causal and non-causal factors is preserved after target adaptation (Fig. \ref{fig:pof}\ak{B}).

\noindent
As Insight \ak{1} motivates that aligning non-causal factors is extremely crucial for preserving the spurious correlation between causal and non-causal factors, a natural question arises, \textit{``How do we enable alignment of non-causal factors between a source and target domain in a SFDA setting?"}

\vspace{1mm}
\noindent
\textbf{Insight 2.\ 
(Style clsf.\ for non-causal factor extraction)
} \textit{
Stylization can facilitate controllable access to the local non-causal
factors. Thus, non-causal factors can be aligned using a subsidiary task of style classification, while respecting the source-free constraints.
}

\noindent
\textbf{Remarks.}
To enable non-causal factor alignment, we propose a subsidiary task of style classification on both the vendor and client-side in a source-free setting. We make use of label-preserving augmentations \cite{piratla2020efficient} (see Suppl.\ for details) to construct novel styles and train a style classifier $f_n$ on a novel style token $z_n \in \mathcal{Z}_n$ in the transformer backbone $f$. Through the subsidiary task of style classification, we propose to extract the non-causal factors and project the local features of the source and target domain into a common feature space. This implicitly aligns the two domains, even without concurrent access to source and target data. Further, it can be easily enabled in a practical source-free setting by sharing only the augmentation information between the vendor and the client, without sharing the data.

\vspace{0.5mm}
\noindent
\textbf{Analysis Experiments.} To analyze the effectiveness of our proposed insights, we examine the effect of non-causal factor alignment on the causal factors. In Fig.\ \ref{fig:pof}\ak{B}, we observe that the $\mathcal{A}$-distance \cite{ben2006analysis} between the causal and non-causal factors remain almost same in both source and target domains, indicating that the spuroius correlation between $S$ and $Z$ is preserved with our approach. For (Fig.\ \ref{fig:pof}\ak{C}), we construct a domain-invariant feature (as a proxy for the causal factors) by taking the mean of class tokens across augmentations. Between the mean class tokens from source and target domains, we compute the $\mathcal{A}$-distance \cite{ben2006analysis}, which measures the separation between the two distributions/domains. We observe a lower $\mathcal{A}$-distance at the class token for our method as compared to the baseline SHOT, indicating that non-causal factor alignment leads to improved causal factor alignment (Insight \ak{1}) and global alignment between the source and target domains (Insight \ak{2}). 

\vspace{2mm}
\subsection{Training Algorithm}
\noindent
We propose a \textbf{C}ausality-enforcing \textbf{S}ource-\textbf{F}ree \textbf{T}ransformer (\textbf{C-SFTrans}) framework which involves a two-stage feature alignment for learning causal representations using vision transformers. 

\subsubsection{Vendor-side source training }
\label{sec:vendor}
The vendor trains C-SFTrans on the source dataset in two steps: \textbf{(a)} \emph{Non-causal factor alignment} using a style classification task, and \textbf{(b)} \emph{Goal task-discriminative feature alignment}, each of which are discussed in detail below.

\noindent
\textbf{a) Non-causal factor alignment.}
\noindent
To align the non-causal factors, we introduce a style classification task that is trained with a novel style token $z_n \in \mathcal{Z}_n$ and updates only non-causal attention heads $h_n^l \in \mathcal{H}$. Here, $\mathcal{H}=\{h_{i}^l\}_{i=1, l=1}^{i=N_h, l=L}$ denotes the set of self-attention heads across all blocks, $N_h$ denotes the number of self-attention heads in each block of the backbone $f$, and $L$ denotes the number of blocks. Each head $h_i^l$ computes self-attention as $h_i^l = \text{Softmax} (\alpha K Q^T) V$. Here, $K=z_j^l W_K$, $Q=z_j^l \cdot W_Q$ and $V=z_j^l W_V$, $\alpha = \frac{1}{\sqrt{d_k}}$, and $d_k$ is the dimension of $z_j^l$. Let ${\mathcal{A}_j^l}_x = h_i^l(\{z_c^l, z_1^l, ..., z_{N_P}^l\}_x)$ where $x$ is the input sample that gets divided into $N_P$ patch tokens and $\mathcal{A}_j^l$ denotes the output of the self-attention head. For the non-causal factor alignment which will update only non-causal attention heads, we first need to select these non-causal attention heads. We choose the heads that give higher importance to the non-causal style features than the causal task-discriminative features. Next, we outline the procedure for non-causal attention head selection.

\begin{table*}[t]
    \centering
    \setlength{\tabcolsep}{1.5pt}
    \caption{Single-Source Domain Adaptation (SSDA) on Office-Home benchmark. SF denotes \textit{source-free} adaptation. ResNet-based methods (top) and Transformer-based methods (bottom). 
    * indicates results taken from \cite{xu2021cdtrans}.
    }
    \label{tab:ssda_oh}
    \vspace{-3mm}
    \resizebox{\linewidth}{!}{%
        \begin{tabular}{lccccccccccccccl}
            \toprule
            \multirow{2}{30pt}{\centering Method} & \multirow{2}{*}{\centering SF} & &\multicolumn{13}{c}{\textbf{Office-Home}} \\
            \cmidrule{4-16}
            && & { {Ar}$\shortrightarrow${Cl}} & { {Ar}$\shortrightarrow${Pr}} & {{Ar}$\shortrightarrow${Rw}} & {{Cl}$\shortrightarrow${Ar}} & {{Cl}$\shortrightarrow${Pr}} & {{Cl}$\shortrightarrow${Rw}} & {{Pr}$\shortrightarrow${Ar}} & {{Pr}$\shortrightarrow${Cl}} & {{Pr}$\shortrightarrow${Rw}} & {{Rw}$\shortrightarrow${Ar}} & {{Rw}$\shortrightarrow${Cl}} & {{Rw}$\shortrightarrow${Pr}} & Avg \\
            \midrule
            ResNet-50 \cite{he2016deep_resnet} & \xmark && 34.9 & 50.0 & 58.0 & 37.4 & 41.9 & 46.2 & 38.5 & 31.2 & 60.4 & 53.9 & 41.2 & 59.9 & 46.1 \\
            ${\text{A}^{2}\text{Net}}$ \cite{A2Net} &\cmark& &  58.4 & 79.0 & {82.4} & 67.5 & 79.3 & 78.9 & {68.0} & 56.2 & 82.9 & {74.1} & 60.5 & 85.0 & 72.8 \\
            GSFDA~\cite{GSFDA} & \cmark && 57.9 & 78.6 & 81.0 & 66.7 & 77.2 & 77.2 & 65.6 & 56.0 & 82.2 & 72.0 & 57.8 & 83.4 & 71.3 \\
            {NRC}~\cite{NRC} &\cmark& &{57.7} 	& {80.3} 	&{82.0} 	&{68.1} 	&{79.8} 	&{78.6} 	&{65.3} 	&{56.4} 	&{83.0} 	&71.0	&{58.6} &{85.6} 	&{72.2} \\
            SHOT~\cite{SHOT} &\cmark& &{57.1} & {78.1} & 81.5 & {68.0} & {78.2} & {78.1} & {67.4} & 54.9 & {82.2} & 73.3 & 58.8 & {84.3} & {71.8} \\
            SHOT++ \cite{SHOT++} & \cmark && 57.9 & 79.7 & 82.5 & {68.5} & 79.6 & 79.3 & {68.5} & 57.0 & {83.0} & 73.7 & 60.7 & 84.9 & 73.0 \\
            \midrule
            TVT \cite{yang2021tvt} & \xmark && 74.8 & 86.8 & 89.4 & 82.7 & 87.9 & 88.2 & 79.8 & 71.9 & 90.1 & 85.4 & 74.6 & 90.5 & 83.5 \\
            SSRT-B \cite{sun2022safe}& \xmark && 75.1 & 88.9 & 91.0 & 85.1 & 88.2 & 89.9 & 85.0 & 74.2 & 91.2 & 85.7 & 78.5 & 91.7 & 85.4 \\
            CDTrans \cite{xu2021cdtrans} & \xmark && 68.8 & 85.0 & 86.9 & 81.5 & 87.1 & 87.3 & 79.6 & 63.3 & 88.2 & 82.0 & 66.0 & 90.6 & 80.5 \\
            \rowcolor{Gray} SHOT-B* & \cmark && 67.1  & 83.5  & 85.5  & 76.6  & 83.4  & 83.7  & 76.3  & 65.3  & 85.3  & 80.4  & 66.7 & 83.4 & {78.1} \improvement{2.5} \\
            \rowcolor{Gray} DIPE \cite{wang2022exploring} & \cmark && 66.0 & 80.6 & 85.6 & 77.1 & 83.5 & 83.4 & 75.3 & 63.3 & 85.1 & 81.6 & 67.7 & 89.6 & 78.2 \improvement{2.4} \\
            \rowcolor{Gray} Mixup \cite{kundu2022balancing} & \cmark && 65.3 & 82.1 & 86.5 & 77.3 & 81.7 & 82.4 & 77.1 & 65.7 & 84.6 & 81.2 & 70.1 & 88.3 & 78.5 \improvement{2.1} \\
            \rowcolor{Gray} \textbf{C-SFTrans (\emph{Ours})} & \cmark && \textbf{70.3} & \textbf{83.9} & \textbf{87.3} & \textbf{80.2} & \textbf{86.9} & \textbf{86.1} & \textbf{78.9} & \textbf{65.0} & \textbf{87.7} & \textbf{82.6} & \textbf{67.9} & \textbf{90.2} & \textbf{80.6} \\
            \bottomrule
            \end{tabular} 
        }
    \vspace{-2mm}
\end{table*}

\vspace{0.5mm}
\noindent
\textbf{Non-causal attention heads selection.} We propose a novel selection criteria to select attention heads based on their contribution towards the causal goal task features and the non-causal style-related features. For this, we first construct a novel \textbf{\emph{Style Characterizing Input} (SCI)} to preserve only style-related features of the input samples. To construct an SCI, we apply a task-destructive transformation (patch shuffling) \cite{mitsuzumi2021generalized}, which keeps the style information intact. Intuitively, we preserve the higher-order statistics of style \cite{chen2020HoMM} by shuffling patches while the class information is destroyed.

\vspace{-1.0mm}
\noindent
We pass both the clean input $x$ and the SCI $x_{SCI}$ as input to each head (Fig.\ \ref{fig:approach}\ak{A}). 
Let ${\mathcal{A}_i^l}_x$  be the output of each head computed using the clean input $x$ computed as follows:
\begin{equation}
    {\mathcal{A}_i^l}_x = h_{i}^l(\{\mathcal{Z}_c^l, \mathcal{Z}_1^l, ..., \mathcal{Z}_{N_P}^l\}_x)
\end{equation}
Let ${\mathcal{A}_i^l}_{x_{SCI}}$ be the output of the heads computed using the input $x_{SCI}$ as follows,
\begin{equation}
    {\mathcal{A}_i^l}_{x_{SCI}} = h_{i}^l(\{\mathcal{Z}_c^l, \mathcal{Z}_1^l, ..., \mathcal{Z}_{N_P}^l\}_{x_{SCI}})
\end{equation}
Let $\beta_1, \beta_2 \in \mathbb{R}^{N_h}$ be the importance weights for the domain and task feature outputs respectively. We constrain these using $\beta_2 = 1 - \beta_1$, which simplifies the optimization. We compute the weighted output as follows,
\begin{equation}
    \mathcal{A}_i^l = \beta_{1_i} \times {\mathcal{A}_i^l}_{x} + \beta_{2_i} \times {\mathcal{A}_i^l}_{x_{SCI}}
\end{equation}
This weighted output $\mathcal{A}_i^l$ is propagated further across the layers. We keep the entire model frozen, training only the two parameters $\beta_1$ and $\beta_2$ for each attention head (Fig.\ \ref{fig:approach}\ak{A}). The following objective is used for attention heads selection,
\begin{equation}
     \min_{\beta_{1}, \beta_{2}} \expectation_{(x_{s}, y_{s}) \in  \mathcal{D}_{s}} [\mathcal{L}_{cls}] 
    \text{ where } \mathcal{L}_{cls} = \mathcal{L}_{ce}(f_g(z_c), y_c)
    \label{eqn:src_clsf_loss}
\end{equation}
See Suppl.\ for more details.
Next, we define the criterion for selecting non-causal heads using the optimized $\beta_1, \beta_2$.

\vspace{1mm}
\noindent
\textbf{Definition 1.\ (Causal Influence Score Criterion)}
\textit{Causal Influence Score (CIS) is computed as $\text{CIS}_i = \beta_{2_i} - \beta_{1_i}$ for an attention head $h_i$. We choose each attention head $h_i$ as non-causal for which the condition $\text{CIS}_i > \tau$ is satisfied. The remaining heads are designated as causal heads.} 

\noindent
\textbf{Remarks.}
Since we train $\beta_1$ and $\beta_2$ with the task classification objective, a higher value of $\beta_2$ indicates that the attention head gives more weightage to the style information and is inherently more suitable for the style classification task. Empirically, we set 30\% heads as non-causal for our experiments, keeping the remaining heads fixed as causal heads for the goal task classification task. While a more sophisticated block-wise strategy could be used, we find this to be insensitive over a wide range of values (see Suppl).

\vspace{0.5mm}
\noindent
\textbf{Style classification task.} Once the causal and non-causal attention heads are chosen, the vendor prepares the augmented datasets $\mathcal{D}_s^{(i)} = \{ (x_s^{[i]}, y_s, y_n) \} \;\forall  \; i \in [N_a]$ by augmenting each source sample $x_s$ (where $N_a$ is the number of augmentations). Here, an augmentation $a_i\!:\! \mathcal{X} \!\to\! \mathcal{X}$ is applied to get $x_s^{[i]} \!=\! a_i(x_s)$. Each input is assigned a style label $y_n\!=\!i$ where $i$ denotes the augmentation label. We use five label-preserving augmentations that simulate novel styles \cite{kundu2021generalize}. Refer to Suppl.\ for more details. 

\vspace{0.5mm}
\noindent
For the non-causal factor learning task (Fig.\ \ref{fig:approach}\ak{B}), we train only the non-causal heads for style classification as follows,

\begin{equation}
    \min_{\theta_{h_{n}}, \theta_{f_{n}}} \expectation_{(x, y_{s}) \in  \cup_{i} \mathcal{D}_{s}^{(i)}} [\mathcal{L}_{style}] \text{ where } \mathcal{L}_{style} = \mathcal{L}_{ce}(f_{n}(z_{n}), y_{n})
    \label{eqn:src_dom_loss}
\end{equation}

\noindent
where $\theta_{h_n}$ contains non-causal heads selected using Def.\ \ak{1}.

\vspace{1mm}
\noindent
\textbf{b) Task-discriminative causal factor alignment.}

\noindent
After one round of style classifier training, we perform the goal task training, where we update only the causal heads (Fig.\ \ref{fig:approach}\ak{B}). The vendor trains the source model consisting of the backbone $f$ and the task classifier $f_g$ with the source labeled dataset $\mathcal{D}_s$ and the task classification loss as follows,
\begin{equation}
    \min_{\theta_f\setminus\theta_{h_n}, \theta_{f_g}} \expectation_{(x_s, y_s) \in  \mathcal{D}_s} [\mathcal{L}_{cls}] 
    \text{ where } \mathcal{L}_{cls} = \mathcal{L}_{ce}(f_g(z_c), y_c)
    \label{eqn:src_clsf_loss_1}
\end{equation}
\noindent
where $z_c$ is the class token, $\theta_f\!\setminus\!\theta_{h_d}$ are the parameters of the backbone excluding the parameters of non-causal heads and $\theta_{f_g}$ are the parameters of task classifier $f_g$.

\noindent
The two steps of non-causal factor alignment and goal task-discriminative feature alignment are performed in alternate iterations, one after the other (see Suppl.\ for details).

\begin{table}[t]
    \centering
    \setlength{\tabcolsep}{4.5pt}
    \caption{Multi-Target Domain Adaptation (MTDA) on Office-Home. SF denotes \textit{source-free} adaptation.} 
    \label{tab:mtda_oh}
    \vspace{-3mm}
    \resizebox{\columnwidth}{!}{%
        \begin{tabular}{lccccccl}
            \toprule
            \multirow{2}{30pt}{\centering Method} & \multirow{2}{*}{\centering SF} &  &\multicolumn{5}{c}{\textbf{Office-Home}} \\
            \cmidrule{4-8}
            && & Ar$\shortrightarrow$ & Cl$\shortrightarrow$ & Pr$\shortrightarrow$ & Rw$\shortrightarrow$ & Avg. \\
            \midrule
            MT-MTDA \cite{nguyen2021unsupervised} & \xmark &  & 64.6 & 66.4 & 59.2 & 67.1 & 64.3 \\
            CDAN+DCL \cite{CDAN} & \xmark & & 63.0 & 66.3 & 60.0 & 67.0 & 64.1 \\
            D-CGCT \cite{D-CGCT} & \xmark & & 70.5 & 71.6 & 66.0 & 71.2 & 69.8 \\
            \midrule
            
            D-CGCT-B \cite{D-CGCT} & \xmark & & 77.0 & 78.5 & 77.9 & 80.9 & 78.6 \\
            \rowcolor{Gray} SHOT-B* & \cmark & & 75.4 & 79.3 & 73.6 & 77.1 & 76.4 \\
            \rowcolor{Gray} \textbf{C-SFTrans-B (\textit{Ours})} & \cmark & & \textbf{77.3} & \textbf{82.9} & \textbf{74.4} & \textbf{76.9} & \textbf{77.8} \color{ForestGreen}\textbf{(+1.4)}\\
            \bottomrule
            \end{tabular} 
        }
\end{table}

\vspace{-2.5mm}
\subsubsection{Client-side target adaptation}
The vendor shares the trained C-SFTrans model with the client for target adaptation. The client performs the non-causal factor alignment in the same way as described earlier by augmenting the target data for style classification. Note that vendor and client may share training and augmentation strategies without sharing the training data \cite{kundu2022balancing}. This step leads to the non-causal factor alignment between source and target domains. For the goal task-discriminative training, the client uses the standard information maximization loss \cite{SHOT} as follows,
\begin{equation}
    \min_{\theta_f\setminus\theta_{h_n}, \theta_{f_g}} \expectation_{\mathcal{D}_t}  [\mathcal{L}_{ent} + \mathcal{L}_{div}] + \min_{\theta_{h_n}, \theta_{f_n}}\expectation_{\cup_{i} \mathcal{D}_t^{(i)}} [\mathcal{L}_{style}]
\end{equation}

\noindent
where $\mathcal{L}_{ent}$, $\mathcal{L}_{div}$ denote entropy and diversity losses, respectively and $\mathcal{L}_{style}$ is supervised CE loss. Note that only original unlabeled target data is used to optimize $\mathcal{L}_{ent}$, $\mathcal{L}_{div}$. The two steps of non-causal factor alignment and task-discriminative causal factor alignment are done one after the other on the client side as well.

\begin{table}[t]
    \centering
    \setlength{\tabcolsep}{10pt}
    \renewcommand{\arraystretch}{0.95}
    \caption{Single-Source Domain Adaptation (SSDA) on Office-31 and VisDA benchmarks. SF denotes \textit{source-free} adaptation.
    ResNet-based methods (top) and Transformer-based methods (bottom). 
    * indicates results taken from \cite{xu2021cdtrans}.
    }
    \label{tab:ssda_o31}
    \vspace{-3mm}
     \resizebox{1\columnwidth}{!}{%
        \begin{tabular}{lcll}
            \toprule
            \textbf{Method} & \textbf{SF} & \begin{tabular}{c}\textbf{Office-31}\end{tabular} & \begin{tabular}{c}\textbf{VisDA}\end{tabular} \\
            \midrule
            ResNet-50 \cite{he2016deep_resnet} & \xmark & 76.1 & 52.4 \\
            NRC~\cite{NRC} & \cmark & 89.4 & 85.9 \\
            SHOT~\cite{SHOT}& \cmark & 88.6 & 82.9 \\
            SHOT++~\cite{SHOT++} & \cmark & 89.2 & 87.3 \\
            
            \midrule
            TVT \cite{yang2021tvt}& \xmark & 93.8 & 83.9 \\
            CGDM-B* \cite{du2021cross} & \xmark & 91.2 & 82.3 \\
            CD-Trans \cite{xu2021cdtrans} & \xmark & 92.6 & 88.4 \\
            SSRT-B \cite{sun2022safe} & \xmark & 93.5 & 88.7 \\
            
            \rowcolor{Gray} SHOT-B* & \cmark & 91.4 \improvement{0.9} & 85.9 \improvement{2.4} \\
            \rowcolor{Gray} DIPE \cite{wang2022exploring} & \cmark & 90.5 \improvement{1.8} & 82.8 \improvement{5.3} \\
            \rowcolor{Gray} Mixup \cite{kundu2022balancing} & \cmark & 91.7 \improvement{0.6} & 86.3 \improvement{2.0} \\
            \rowcolor{Gray} \textbf{C-SFTrans (\textit{Ours})} & \cmark & \textbf{92.3} & \textbf{88.3} \\

            \bottomrule
            \end{tabular} 
         }
         \vspace{-3mm}
\end{table}

\begin{table*}[t]
\centering
\setlength{\tabcolsep}{2.2pt}
\caption{Single-Source Domain Adaptation (SSDA) on the DomainNet benchmark. * indicates results taken from \cite{sun2022safe}.}
\label{tab:ssda_dnet}
\vspace{-4mm}
\resizebox{\textwidth}{!}{%
\begin{tabular}[t]{ccc}
        \begin{tabular}[t]{|C{2cm}|ccccccc|}
            \hline
            \begin{tabular}{c} \textbf{MDD+} \\  \textbf{SCDA} \cite{MDD} \end{tabular} & clp & inf & pnt & qdr & rel & skt & Avg. \\
            \hline
            clp  & -    & 20.4 & 43.3 & 15.2 & 59.3 & 46.5 & 36.9 \\
            inf  & 32.7 & -    & 34.5 & 6.3  & 47.6 & 29.2 & 30.1 \\
            pnt  & 46.4 & 19.9 & -    & 8.1  & 58.8 & 42.9 & 35.2 \\
            qdr  & 31.1 & 6.6  & 18.0 & -    & 28.8 & 22.0 & 21.3 \\
            rel  & 55.5 & 23.7 & 52.9 & 9.5  & -    & 45.2 & 37.4 \\
            skt  & 55.8 & 20.1 & 46.5 & 15.0 & 56.7 & -    & 38.8 \\
            Avg. & 44.3 & 18.1 & 39.0 & 10.8 & 50.2 & 37.2 & \cellcolor{Gray} \textbf{33.3} \\
            \hline
        \end{tabular} &
        
        \begin{tabular}[t]{|C{2cm}|ccccccc|}
            \hline
            \begin{tabular}{c} \textbf{DeiT-B} \\  \cite{touvron2021training} \end{tabular} & clp & inf & pnt & qdr & rel & skt & Avg. \\
            \hline
            clp  & -    & 24.3 & 49.6 & 15.8 & 65.3 & 52.1 & 41.4 \\
            inf  & 45.9 & -    & 45.9 & 6.7  & 61.4 & 39.5 & 39.9 \\
            pnt  & 53.2 & 23.8 & -    & 6.5  & 66.4 & 44.7 & 38.9 \\
            qdr  & 31.9 & 6.8  & 15.4 & -    & 23.4 & 20.6 & 19.6 \\
            rel  & 59.0 & 25.8 & 56.3 & 9.16 & -    & 44.8 & 39.0 \\
            skt  & 60.6 & 20.6 & 48.4 & 16.5 & 61.2 & -    & 41.5 \\
            Avg. & 50.1 & 20.3 & 43.1 & 10.9 & 55.5 & 40.3 & \cellcolor{Gray} \textbf{36.7} \\
            \hline
        \end{tabular} &
        
        \begin{tabular}[t]{|C{2cm}|ccccccc|}
            \hline
            \begin{tabular}{c} \textbf{SHOT-B} \\  \cite{SHOT} \end{tabular} & clp & inf & pnt & qdr & rel & skt & Avg. \\
            \hline
            clp  & -    & 45.9 & 49.7 & 16.5 & 65.4 & 53.2 & 46.1 \\
            inf  & 46.4 & -    & 45.9 & 7.4  & 60.6 & 40.1 & 40.1 \\
            pnt  & 54.6 & 25.7 & -    & 8.1  & 66.3 & 49.0 & 40.7 \\
            qdr  & 33.3 & 6.8  & 15.5 & -    & 23.8 & 24.0 & 20.7 \\
            rel  & 59.3 & 28.1 & 57.4 & 9.0  & -    & 47.3 & 40.2 \\
            skt  & 64.0 & 26.5 & 55.0 & 18.2 & 63.8 & -    & 45.5 \\
            Avg. & 51.5 & 26.6 & 44.7 & 11.8 & 56.0 & 42.7 & \cellcolor{Gray} \textbf{38.9} \\
            \hline
        \end{tabular} \\
        \vspace{-2mm}
        \begin{tabular}[t]{|C{2cm}|ccccccc|}
            \hline
            \begin{tabular}{c} \textbf{CDTrans$^*$} \\  \cite{xu2021cdtrans} \end{tabular} & clp & inf & pnt & qdr & rel & skt & Avg. \\
            \hline
            clp  & -    & 27.9 & 57.6 & 27.9 & 73.0 & 58.8 & 49.0 \\
            inf  & 58.6 & -    & 53.4 & 9.6  & 71.1 & 47.6 & 48.1 \\
            pnt  & 60.7 & 24.0 & -    & 13.0 & 69.8 & 49.6 & 43.4 \\
            qdr  & 2.9  & 0.4  & 0.3  & -    & 0.7  & 4.7  & 1.8  \\
            rel  & 49.3 & 18.7 & 47.8 & 9.4  & -    & 33.5 & 31.7 \\
            skt  & 66.8 & 23.7 & 54.6 & 27.5 & 68.0 & -    & 48.1 \\
            Avg. & 47.7 & 18.9 & 42.7 & 17.5 & 56.5 & 38.8 & \cellcolor{Gray} \textbf{37.0} \\
            \hline
        \end{tabular} &
        
        \begin{tabular}[t]{|C{2cm}|ccccccc|}
            \hline
            \begin{tabular}{c} \textbf{SSRT-B$^*$} \\  \cite{sun2022safe} \end{tabular} & clp & inf & pnt & qdr & rel & skt & Avg. \\
            \hline
            clp  & -    & 33.8 & 60.2 & 19.4 & 75.8 & 59.8 & 49.8 \\
            inf  & 55.5 & -    & 54.0 & 9.0  & 68.2 & 44.7 & 46.3 \\
            pnt  & 61.7 & 28.5 & -    & 8.4  & 71.4 & 55.2 & 45.0 \\
            qdr  & 42.5 & 8.8  & 24.2 & -    & 37.6 & 33.6 & 29.3 \\
            rel  & 69.9 & 37.1 & 66.0 & 10.1 & -    & 58.9 & 48.4 \\
            skt  & 70.6 & 32.8 & 62.2 & 21.7 & 73.2 & -    & 52.1 \\
            Avg. & 60.0 & 28.2 & 53.3 & 13.7 & 65.3 & 50.4 & \cellcolor{Gray} \textbf{45.2} \\
            \hline
        \end{tabular} &
        
        \begin{tabular}[t]{|C{2cm}|ccccccc|}
            \hline
            \begin{tabular}{c} \textbf{C-SFTrans} \\  \textbf{(Ours)} \end{tabular} & clp & inf & pnt & qdr & rel & skt & Avg. \\
            \hline
            clp  & -    & 26.6 & 53.6 & 23.6 & 71.4 & 54.6 & 46.0 \\
            inf  & 55.9 & -    & 51.7 & 11.4 & 69.6 & 46.0 & 46.9 \\
            pnt  & 60.0 & 25.2 & -    & 14.3 & 71.2 & 51.1 & 44.4 \\
            qdr  & 43.2 & 8.2  & 17.4 & -    & 40.2 & 28.8 & 27.5 \\
            rel  & 60.4 & 28.1 & 56.5 & 12.2 & -    & 49.8 & 41.4 \\
            skt  & 66.7 & 26.5 & 56.2 & 25.1 & 71.0 & -    & 49.1 \\
            Avg. & 57.2 & 22.9 & 47.1 & 17.3 & 64.7 & 46.1 & \cellcolor{Gray} \textbf{42.5} \\
            \hline
        \end{tabular} \\
        \vspace{-3mm}
\end{tabular}}
\end{table*}

\section{Experiments}
In this section, we evaluate our proposed approach by comparing with existing works on several benchmarks and analyze the significance of each component of the approach.
\noindent
\textbf{Datasets.} We evaluate our approach on four existing standard object classification benchmarks for Domain Adaptation: OfficeHome, Office-31, VisDA, and DomainNet. The \textbf{Office-Home} dataset \cite{office-home} contains images from 65 categories of objects found in everyday home and office environments. The images are grouped into four domains - Art \textbf{(Ar)}, Clipart \textbf{(Cl)}, Product \textbf{(Pr)} and Real World \textbf{(Rw)}. The \textbf{Office-31 (Office)} dataset \cite{office-31} consists of images from three domains - Amazon \textbf{(A)}, DSLR \textbf{(D)}, and Webcam \textbf{(W)}. The three domains contain images from 31 classes of objects that are found in a typical office setting. The \textbf{VisDA} \cite{visda} dataset is a large-scale synthetic-to-real benchmark with images from 12 categories. \textbf{DomainNet} \cite{M3SDA} is the largest and the most challenging dataset among the four standard benchmarks due to severe class imbalance and diversity of domains. It contains 345 categories of objects from six domains - Clipart \textbf{(clp)}, Infograph \textbf{(inf)}, Painting \textbf{(pnt)}, Quickdraw \textbf{(qdr)}, Real \textbf{(rel)}, Sketch \textbf{(skt)}.

\noindent
\textbf{Implementation details.} To ensure fair comparisons, we make use of DeiT-Base \cite{touvron2021training} with patch size $16 \times 16$ and follow the experimental setup outlined in CDTrans \cite{xu2021cdtrans}. We use Stochastic Gradient Descent (SGD) with a weight decay ratio of $1 \times 10^{-4}$, and a momentum of $0.9$ for the training process. Refer to Suppl.\ for more implementation details.

\begin{table}[H]
    \centering
    \setlength{\tabcolsep}{5.5pt}
    \vspace{-3mm}
    \caption{Ablation study for various stages of training on VisDA Single-Source Domain Adaptation (SSDA) benchmark.
    } 
    \label{tab:ablation}
    \vspace{-3mm}
    \resizebox{\columnwidth}{!}{%
        \begin{tabular}{lcccl}
        \toprule
         Training Phase & Method & \begin{tabular}{c} Goal \\ Task \end{tabular} & \begin{tabular}{c} Style \\ Task \end{tabular} & Avg. \\
         \midrule
        \multirow{3}{*}{\begin{tabular}{c} Source-Side \end{tabular}} & Source-Only & \cmark & \xmark &  65.1 \\
                                     & \cellcolor{Gray} & \cellcolor{Gray}\xmark & \cellcolor{Gray}\cmark & \cellcolor{Gray}66.2 \color{ForestGreen}\textbf{(+1.1)} \\
                                     & \multirow{-2}{*}{\cellcolor{Gray}Ours} & \cellcolor{Gray}\cmark & \cellcolor{Gray}\cmark & \cellcolor{Gray}68.8 \color{ForestGreen}\textbf{(+3.7)} \\
        \hline
         \multirow{3}{*}{\begin{tabular}{c}Target-Side \end{tabular}} \rule[10pt]{0pt}{0pt} & SHOT-B & \cmark & \xmark & 85.9 \\
                                     \rule[10pt]{0pt}{0pt} & \cellcolor{Gray}& \cellcolor{Gray}\xmark & \cellcolor{Gray}\cmark & \cellcolor{Gray}87.5 \color{ForestGreen}\textbf{(+1.6)}\\
                                     & \multirow{-2}{*}{\cellcolor{Gray}Ours} & \cellcolor{Gray}\cmark & \cellcolor{Gray}\cmark & \cellcolor{Gray}\textbf{88.3} \color{ForestGreen}\textbf{(+2.4)}\\          
        \bottomrule
        \end{tabular}
        }
        \vspace{-2mm}
\end{table}

\subsection{Comparison with prior arts}
\label{sota_comparison}

\noindent
\textbf{a) Single-Source Domain Adaptation (SSDA).}
\label{ssda}
We provide comparisons between our proposed method, C-SFTrans, and earlier SSDA works in Tables \ref{tab:ssda_oh} and \ref{tab:ssda_o31}. Our method provides the best performance among source-free works for the three standard DA benchmarks. On Office-Home, C-SFTrans outperforms the transformer based source-free prior work SHOT-B* by $2.5\%$ and shows competitive performance \wrt the non-source-free method CDTrans \cite{xu2021cdtrans}. On the Office-31 benchmark (Table \ref{tab:ssda_o31}), our technique outperforms  the source-free SHOT-B* by $0.9\%$ and achieves competitive performance when compared to non-source-free works.  Table \ref{tab:ssda_o31} also demonstrates that our method shows $2.4\%$ improvement over SHOT-B* and is on par with the non-source-free methods CDTrans and SSRT \cite{sun2022safe} on the larger and more challenging VisDA benchmark. We also achieve a significant improvement of 3.6\% on the DomainNet benchmark (Table \ref{tab:ssda_dnet}) over SHOT-B baseline. \par

\vspace{1mm}
\noindent
\textbf{b) Multi Target Domain Adaptation (MTDA).}
\label{mtda}
In Table \ref{tab:mtda_oh}, we compare our proposed framework, C-SFTrans, with existing works on multi-target domain adaption on the OfficeHome dataset. Our method achieves a $1.4\%$ improvement over the source-free baseline (SHOT-B) and is comparable to the non-source-free method D-CGCT \cite{D-CGCT} despite using a pure transformer backbone while the latter uses a hybrid convolution-transformer feature extractor.

\subsection{Analysis}
\label{analysis}
\noindent
We perform a thorough ablation study of our proposed approach and analyze the contribution of each component of our approach in Table \ref{tab:ablation}.

\vspace{1mm}
\noindent
\textbf{a) Effect of non-causal factor classification.}
In Table \ref{tab:ablation}, we study the effect of non-causal factor classification on the goal task performance using a subsidiary style classification task. We first train only the style task while keeping the goal task classifier $f_g$ and causal parameters $\theta_{f}\!\setminus\!\theta_{h_n}$ fixed.Here, we observe a significant improvement of 1.1\% on the source-side and 1.6\% on the target-side. This validates Insight \ak{2} since the non-causal style classification task improves the causal factor alignment, thereby improving the goal task performance. 

\vspace{1mm}
\noindent
\textbf{b) Effect of both goal and non-causal tasks.} 
Next, in Table \ref{tab:ablation}, we use both goal task and non-causal style task alternately as proposed in Sec.\ \ref{sec:approach}. As per Insight \ak{1}, this should improve the alignment between source and target causal factors, and result in optimal clustering of task-related features. The alternate training of style and goal task yields an overall improvement of 3.7\% on source-side and 2.4\% on target-side, which validates Insight \ak{1}.

\vspace{1mm}
\noindent

\begin{table}[t]
    \centering
    \setlength{\tabcolsep}{2.2pt}
    \caption{Single-Source Domain Adaptation (SSDA) on Office-Home with ViT-B Backbone pre-trained on the ImageNet-21K dataset (\emph{bottom}), and the DeiT-S backbone pre-trained on ImageNet-1K dataset (\emph{top}). SF indicates \textit{source-free} adaptation.} 
    \label{tab:vit-b}
    \vspace{-3mm}
    \resizebox{\columnwidth}{!}{%
        \setlength{\extrarowheight}{1pt}
        \begin{tabular}{lccccccl}
            \toprule
            \multirow{2}{30pt}{\centering Method} & \multirow{2}{*}{\centering SF} &  &\multicolumn{5}{c}{\textbf{Office-Home}} \\
            \cmidrule{4-8}
            && & Ar$\shortrightarrow$Cl & Cl$\shortrightarrow$Pr & Pr$\shortrightarrow$Rw & Rw$\shortrightarrow$Ar & Avg. \\
            \midrule
            CDTrans-S \cite{xu2021cdtrans} & \xmark && 60.7 & 75.6 & 84.4 & 77.0 & 74.4 \\
            \rowcolor{Gray} SHOT-S & \cmark && 56.3 & 73.7 & 81.3 & 76.7 & 72.0 \\
            \rowcolor{Gray} \textbf{C-SFTrans-S} & \cmark &&  \textbf{63.3} & \textbf{79.7} & \textbf{83.0} & \textbf{76.8} & \textbf{75.7} \textbf{\color{ForestGreen}{(+3.7)}}\\
            \midrule
            SSRT-B \cite{sun2022safe} & \xmark & & 75.2 & 88.3 & 91.3 & 85.7 & 85.1 \\
            \rowcolor{Gray} SHOT-B & \cmark & & 69.1 & 85.3 & 88.1 & 83.9 & 81.6 \\
            \rowcolor{Gray} \textbf{C-SFTrans-B (\textit{Ours})} & \cmark & & \textbf{73.1} & \textbf{87.3}	& \textbf{83.5} & \textbf{88.1} & \textbf{83.0} \textbf{\color{ForestGreen}{(+1.4)}}\\
            \bottomrule
            \end{tabular} 
        }
        \vspace{-2.5mm}
\end{table}

\noindent
\textbf{c) Comparisons with different backbones.} In Table \ref{tab:vit-b}, we provide results for our approach with the ViT-Base backbone pre-trained on the ImageNet-21K dataset, and the DeiT-S backbone pre-trained on the ImageNet-1K dataset. We observe that our method achieves a 1.4\% improvement over SHOT-B with the ViT-B backbone, and a 3.7\% improvement over SHOT-S with the DeiT-S backbone.

\section{Conclusion}
In this work, we study the concepts of source-free domain-adaptation from the perspective of causality. We conjecture that the spurious correlation among causal and non-causal factors are crucial to preserve in the target domain to improve the adaptation performance. Hence, we provide insights showing that the disentangling and aligning non-causal factors positively influence the alignment of causal factors in SFDA. Further, we first investigate the behavior of vision transformers in SFDA and propose a novel Causality-enforcing Source-free Transformer (C-SFTrans) architecture for non-causal factor alignment. Based on our insights, we introduce a non-causal factor classification task to align non-causal factors. We also propose a novel Causal Influence Score criterion to improve the training. The proposed approach leads to improved task-discriminative causal factor alignment and outperforms the prior works on DA benchmarks of single-source and multi-target SFDA. \textbf{Acknowledgements.} Sunandini Sanyal was supported by the Prime Minister's Research Fellowship, Govt of India.

\appendix
\twocolumn[
    \begin{@twocolumnfalse}
    \begin{center}
        \textbf{\Large Supplementary Material}
        \vspace{1cm}
    \end{center}
    \end{@twocolumnfalse}
]

\noindent
The supplementary material provides further details of the proposed approach, additional quantitative results, ablations, and implementation details. We have released our code on our project page: \url{https://val.cds.iisc.ac.in/C-SFTrans/}. The remainder of the supplementary material is organized as follows:

\begin{itemize}
    
    \item Section \ref{sec:approach_suppl}: Proposed Approach (Table \ref{sup:tab:notations}, Algorithm \ref{algo:overall})
    
    \item Section \ref{sec:impl}: Implementation Details
    \begin{itemize}
        \item Datasets (Section \ref{sub:datasets})
        \item Style augmentations (Section \ref{sub:style})
        \item Experimental Settings (Section \ref{sub:exp_set})
    \end{itemize}
    
    \item Section \ref{sec:addn_comp}: Additional Comparisons (Tables \ref{tab:ssda_dnet_full})
    
    \item Section \ref{sec:abl_ta}: Ablations on target-side goal task training (Tables \ref{tab:sensitivity_analysis}, \ref{tab:ablation_loss}, and \ref{tab:dom_heads})
    
\end{itemize}

\section{Proposed Approach}
\label{sec:approach_suppl}

\noindent
We summarize all the notations used in the paper in Table \ref{sup:tab:notations}. The notations are grouped into the following 6 categories - models, transformers, datasets, spaces, losses, and criteria. Our proposed method has been outlined in Algorithm \ref{algo:overall}

\begin{table}[ht]
    \centering
    \caption{\textbf{Notation Table}}
    \label{sup:tab:notations}
    \vspace{-3mm}
    \setlength{\tabcolsep}{11pt}
    \resizebox{0.9\columnwidth}{!}{%
        \begin{tabular}{lcl}
        \toprule
        
         & \multirow{1}{*}{Symbol} & \multirow{1}{*}{Description} \\
         \midrule

         \multirow{3}{*}{\rotatebox[origin=c]{90}{Models}} & $f$ & Backbone feature extractor  \\
         & $f_g$ & Goal task classifier \\
         & $f_n$ & Style classifier \\
         \midrule

        \multirow{9}{*}{\rotatebox[origin=c]{90}{Transformers}} 
         & $z_c$ & Class token of last layer \\
         
         & $z_n$ & Style token of last layer\\
        
         & $N_P$ & Number of patch tokens \\
         & $h_n^l$ & Non-causal heads of layer $l$\\
         & $h^l$ & All attention-heads of layer $l$\\
         & $h^l \setminus h_n^l$ & Causal heads of layer $l$\\
         & $W_K$ & Key weights\\
         & $W_Q$ & Query weights\\
         & $W_V$ & Value weights \\

        \midrule
        \multirow{9}{*}{\rotatebox[origin=c]{90}{Datasets}} & $\mathcal{D}_s$ & Labeled source dataset  \\
         & $\mathcal{D}_t$ & Unlabeled target dataset  \\
         & $a_i$ & Augmentation function $i$ \\
         & $\mathcal{D}_s^{[i]}$ & $i^\text{th}$ augmented source dataset \\
         & $\mathcal{D}_t^{[i]}$ & $i^\text{th}$ augmented target dataset \\
         & $(x_s, y_{s})$ & Labeled source sample  \\
         & $(x_{s}^{[i]}, y_s, y_d)$ & Augmented source sample  \\
         & $x_t$ & Unlabeled target sample  \\
         & $(x_{t}^{[i]}, y_d)$ & Target augmented sample \\
         & $x$ & Clean input sample \\
         & $x_{SCI}$ & Style Characterizing Input \\
         
         \midrule 
        \multirow{5}{*}{\rotatebox[origin=c]{90}{Spaces}} 
        & $\mathcal{X}$ & Input space \\
        & $\mathcal{C}_g$ & Label set for goal task \\
        & $\mathcal{Z}_c$ & Class token feature space\\
        & $\mathcal{Z}_n$ & Style token feature space\\
        & $\mathcal{Z}_1,\dots, \mathcal{Z}_{N_P}$ & Patch tokens \\

        \midrule
        \multirow{4}{*}{\rotatebox[origin=c]{90}{ Losses}} & $\mathcal{L}_{style}$ & Style Classification loss  \\
         & $\mathcal{L}_{cls}$ & Task Classification loss  \\
         & $\mathcal{L}_{ent}$ & Entropy loss  \\
         & $\mathcal{L}_{div}$ & Diversity loss \\

         \midrule
        \multirow{4}{*}{\rotatebox[origin=c]{90}{ Criterion}}
         & $\beta_{1_{i}}$ & Importance weight for style feature \\
         & $\beta_{2_{i}}$ & Importance weight for task feature \\
         & $CIS_i$ & Causal Influence Score for head $i$ \\
         & $\tau$ & Threshold  \\
        
        \bottomrule
        \end{tabular}
        }
\end{table}

\vspace{1mm}
\noindent
\textbf{Target adaptation losses.}
We use the Information Maximization loss \cite{SHOT} that consists of entropy loss $\mathcal{L}_{ent}$ and diversity loss $\mathcal{L}_{div}$. 
\begin{equation}
\label{sup:eqn:loss_im}
    \mathcal{L}_{ent} = -\expectation_{x_t \in \mathcal{X}} \sum_{k=1}^{K} \delta_{k}(f_g(z_c)) \log \delta_{k}(f_g(z_c))
\end{equation}
\begin{equation}
\label{sup:eqn:loss_div}
    \mathcal{L}_{div} = \sum_{k=1}^{K} \hat{p_k} \log \hat{p_k}
     = KL (\hat{p}, \frac{1}{K} 1_K) - \log K
\end{equation}
\noindent
where $\delta_{k}(b) = \frac{\exp(b_k)}{\sum_i \exp(b_i)}$ is the $k^{\text{th}}$ element of softmax output of $b \in \mathbb{R}^K$. The entropy loss $\mathcal{L}_{ent}$ ensures that the model predicts more confidently for a particular label and the diversity loss $\mathcal{L}_{div}$ ensures that the predictions are well-balanced across different classes. We optimize all parameters of the transformer backbone $h$, except the non-causal heads $h_n^l$.

\begin{equation}
    \min_{h^l \setminus h_n^l, f_g} \expectation_{\mathcal{D}_t}  [\mathcal{L}_{ent} + \mathcal{L}_{div}] 
\end{equation}
\noindent
\textbf{Pseudo-labeling.}
We use the clustering method of SHOT \cite{SHOT} to obtain pseudo-labels. At first, the centroid of each class is calculated using the following,
\begin{equation}
    c_k = \frac{\sum_{x_t \in \mathcal{X}} \delta_k(f_g(z_c)) z_c}{\sum_{x_t \in \mathcal{X}} \delta_k (f_g(z_c))}
\end{equation}
\noindent
The closest centroid is chosen as the pseudo-label for each sample using the following cosine distance formulation,
\begin{equation}
    \hat{y_c} = \argmin_{k}D_{c}(z_{c}, c_{k})
\end{equation}
where $D_c$ denotes the cosine-distance in the class-token feature space $\mathcal{Z}_c$ between a centroid $c_k$ and the input sample features $z_c$. In successive iterations, the centroids keep updating and the pseudo-labels get updates with respect to the new centroids.

\noindent
\textbf{Attention heads in vision transformers.}
A ViT takes an image $x$ as input of size $H \times W \times C$ and divides it into $N_P$ patches of size $(P,P)$ each. The total number of patches are $N_P = \frac{H \times W}{P^2}$. In every layer, $l$, a head $h_i^l$ takes the patches as input and transforms a patch into $K, Q, V$ using the weights $W_K, W_Q, W_V$, respectively. The self-attention \cite{vaswani2017attention} is computed as follows,
 \begin{equation}
     h_i^l =  \text{Softmax}\left( \frac{QK^T}{\sqrt{d_k}} \right) V
 \end{equation}
where $d_k$ represents the dimension of the keys/queries.

\begin{table*}[t]
\centering
\setlength{\tabcolsep}{2.2pt}
\caption{Single-Source Domain Adaptation (SSDA) results on the DomainNet dataset. * indicates results taken from \cite{sun2022safe}.}
\vspace{-5mm}
\label{tab:ssda_dnet_full}
\resizebox{\textwidth}{!}{
\begin{tabular}[t]{ccc}
    
        \begin{tabular}[t]{|C{2cm}|ccccccc|}
            \hline
            \begin{tabular}{c} \textbf{ResNet-} \\ \textbf{101} \cite{he2016deep_resnet} \end{tabular} & clp & inf & pnt & qdr & rel & skt & Avg. \\
            \hline
            clp  & -    & 19.3 & 37.5 & 11.1 & 52.2 & 41.0 & 32.2 \\
            inf  & 30.2 & -    & 31.2 & 3.6  & 44.0 & 27.9 & 27.4 \\
            pnt  & 39.6 & 18.7 & -    & 4.9  & 54.5 & 36.3 & 30.8 \\
            qdr  & 7.0  & 0.9  & 1.4  & -    & 4.1  & 8.3  & 4.3  \\
            rel  & 48.4 & 22.2 & 49.4 & 6.4  & -    & 38.8 & 33.0 \\
            skt  & 46.9 & 15.4 & 37.0 & 10.9 & 47.0 & -    & 31.4 \\
            Avg. & 34.4 & 15.3 & 31.3 & 7.4  & 40.4 & 30.5 & \cellcolor{Gray} \textbf{26.6} \\
            \hline
        \end{tabular} &
        \begin{tabular}[t]{|C{2cm}|ccccccc|}
            \hline
            \begin{tabular}{c} \textbf{CDAN} \\ \cite{CDAN} \end{tabular} & clp & inf & pnt & qdr & rel & skt & Avg. \\
            \hline
            clp  & -    & 20.4 & 36.6 & 9.0  & 50.7 & 42.3 & 31.8 \\
            inf  & 27.5 & -    & 25.7 & 1.8  & 34.7 & 20.1 & 22.0 \\
            pnt  & 42.6 & 20.0 & -    & 2.5  & 55.6 & 38.5 & 31.8 \\
            qdr  & 21.0 & 4.5  & 8.1  & -    & 14.3 & 15.7 & 12.7 \\
            rel  & 51.9 & 23.3 & 50.4 & 5.4  & -    & 41.4 & 34.5 \\
            skt  & 50.8 & 20.3 & 43.0 & 2.9  & 50.8 & -    & 33.6 \\
            Avg. & 38.8 & 17.7 & 32.8 & \textcolor{white}{0}4.3  & 41.2 & 31.6 & \cellcolor{Gray} \textbf{27.7} \\
            \hline
        \end{tabular} &
        \begin{tabular}[t]{|C{2cm}|ccccccc|}
            \hline
            \begin{tabular}{c} \textbf{MIMFTL} \\  \cite{MIMFTL} \end{tabular} & clp & inf & pnt & qdr & rel & skt & Avg. \\
            \hline
            clp  & -    & 15.1 & 35.6 & 10.7 & 51.5 & 43.1 & 31.2 \\
            inf  & 32.1 & -    & 31.0 & 2.9  & 48.5 & 31.0 & 29.1 \\
            pnt  & 40.1 & 14.7 & -    & 4.2  & 55.4 & 36.8 & 30.2 \\
            qdr  & 18.8 & 3.1  & 5.0  & -    & 16.0 & 13.8 & 11.3 \\
            rel  & 48.5 & 19.0 & 47.6 & 5.8  & -    & 39.4 & 32.1 \\
            skt  & 51.7 & 16.5 & 40.3 & 12.3 & 53.5 & -    & 34.9 \\
            Avg. & 38.2 & 13.7 & 31.9 & 7.2  & 45.0 & 32.8 & \cellcolor{Gray} \textbf{28.1} \\
            \hline
        \end{tabular} \\
        \begin{tabular}[t]{|C{2cm}|ccccccc|}
            \hline
            \begin{tabular}{c} \textbf{MDD+} \\  \textbf{SCDA} \cite{MDD} \end{tabular} & clp & inf & pnt & qdr & rel & skt & Avg. \\
            \hline
            clp  & -    & 20.4 & 43.3 & 15.2 & 59.3 & 46.5 & 36.9 \\
            inf  & 32.7 & -    & 34.5 & 6.3  & 47.6 & 29.2 & 30.1 \\
            pnt  & 46.4 & 19.9 & -    & 8.1  & 58.8 & 42.9 & 35.2 \\
            qdr  & 31.1 & 6.6  & 18.0 & -    & 28.8 & 22.0 & 21.3 \\
            rel  & 55.5 & 23.7 & 52.9 & 9.5  & -    & 45.2 & 37.4 \\
            skt  & 55.8 & 20.1 & 46.5 & 15.0 & 56.7 & -    & 38.8 \\
            Avg. & 44.3 & 18.1 & 39.0 & 10.8 & 50.2 & 37.2 & \cellcolor{Gray} \textbf{33.3} \\
            \hline
        \end{tabular} &
        \begin{tabular}[t]{|C{2cm}|ccccccc|}
            \hline
            \begin{tabular}{c} \textbf{DeiT-B} \\  \cite{touvron2021training} \end{tabular} & clp & inf & pnt & qdr & rel & skt & Avg. \\
            \hline
            clp  & -    & 24.3 & 49.6 & 15.8 & 65.3 & 52.1 & 41.4 \\
            inf  & 45.9 & -    & 45.9 & 6.7  & 61.4 & 39.5 & 39.9 \\
            pnt  & 53.2 & 23.8 & -    & 6.5  & 66.4 & 44.7 & 38.9 \\
            qdr  & 31.9 & 6.8  & 15.4 & -    & 23.4 & 20.6 & 19.6 \\
            rel  & 59.0 & 25.8 & 56.3 & 9.16 & -    & 44.8 & 39.0 \\
            skt  & 60.6 & 20.6 & 48.4 & 16.5 & 61.2 & -    & 41.5 \\
            Avg. & 50.1 & 20.3 & 43.1 & 10.9 & 55.5 & 40.3 & \cellcolor{Gray} \textbf{36.7} \\
            \hline
        \end{tabular} &
        \begin{tabular}[t]{|C{2cm}|ccccccc|}
            \hline
            \begin{tabular}{c} \textbf{SHOT-B} \\  \cite{SHOT} \end{tabular} & clp & inf & pnt & qdr & rel & skt & Avg. \\
            \hline
            clp  & -    & 27.0 & 49.7 & 16.5 & 65.4 & 53.2 & 46.1 \\
            inf  & 46.4 & -    & 45.9 & 7.4  & 60.6 & 40.1 & 40.1 \\
            pnt  & 54.6 & 25.7 & -    & 8.1  & 66.3 & 49.0 & 40.7 \\
            qdr  & 33.3 & 6.8  & 15.5 & -    & 23.8 & 24.0 & 20.7 \\
            rel  & 59.3 & 28.1 & 57.4 & 9.0  & -    & 47.3 & 40.2 \\
            skt  & 64.0 & 26.5 & 55.0 & 18.2 & 63.8 & -    & 45.5 \\
            Avg. & 51.5 & 26.6 & 44.7 & 11.8 & 56.0 & 42.7 & \cellcolor{Gray} \textbf{38.9} \\
            \hline
        \end{tabular} \\
        \begin{tabular}[t]{|C{2cm}|ccccccc|}
            \hline
            \begin{tabular}{c} \textbf{CDTrans$^*$} \\  \cite{xu2021cdtrans} \end{tabular} & clp & inf & pnt & qdr & rel & skt & Avg. \\
            \hline
            clp  & -    & 27.9 & 57.6 & 27.9 & 73.0 & 58.8 & 49.0 \\
            inf  & 58.6 & -    & 53.4 & 9.6  & 71.1 & 47.6 & 48.1 \\
            pnt  & 60.7 & 24.0 & -    & 13.0 & 69.8 & 49.6 & 43.4 \\
            qdr  & 2.9  & 0.4  & 0.3  & -    & 0.7  & 4.7  & 1.8  \\
            rel  & 49.3 & 18.7 & 47.8 & 9.4  & -    & 33.5 & 31.7 \\
            skt  & 66.8 & 23.7 & 54.6 & 27.5 & 68.0 & -    & 48.1 \\
            Avg. & 47.7 & 18.9 & 42.7 & 17.5 & 56.5 & 38.8 & \cellcolor{Gray} \textbf{37.0} \\
            \hline
        \end{tabular} &
        \begin{tabular}[t]{|C{2cm}|ccccccc|}
            \hline
            \begin{tabular}{c} \textbf{SSRT-B$^*$} \\  \cite{sun2022safe} \end{tabular} & clp & inf & pnt & qdr & rel & skt & Avg. \\
            \hline
            clp  & -    & 33.8 & 60.2 & 19.4 & 75.8 & 59.8 & 49.8 \\
            inf  & 55.5 & -    & 54.0 & 9.0  & 68.2 & 44.7 & 46.3 \\
            pnt  & 61.7 & 28.5 & -    & 8.4  & 71.4 & 55.2 & 45.0 \\
            qdr  & 42.5 & 8.8  & 24.2 & -    & 37.6 & 33.6 & 29.3 \\
            rel  & 69.9 & 37.1 & 66.0 & 10.1 & -    & 58.9 & 48.4 \\
            skt  & 70.6 & 32.8 & 62.2 & 21.7 & 73.2 & -    & 52.1 \\
            Avg. & 60.0 & 28.2 & 53.3 & 13.7 & 65.3 & 50.4 & \cellcolor{Gray} \textbf{45.2} \\
            \hline
        \end{tabular} &
        \begin{tabular}[t]{|C{2cm}|ccccccc|}
            \hline
            \begin{tabular}{c} \textbf{C-SFTrans} \\  \textbf{(Ours)} \end{tabular} & clp & inf & pnt & qdr & rel & skt & Avg. \\
            \hline
            clp  & -    & 26.6 & 53.6 & 23.6 & 71.4 & 54.6 & 46.0 \\
            inf  & 55.9 & -    & 51.7 & 11.4 & 69.6 & 46.0 & 46.9 \\
            pnt  & 60.0 & 25.2 & -    & 14.3 & 71.2 & 51.1 & 44.4 \\
            qdr  & 43.2 & 8.2  & 17.4 & -    & 40.2 & 28.8 & 27.5 \\
            rel  & 60.4 & 28.1 & 56.5 & 12.2 & -    & 49.8 & 41.4 \\
            skt  & 66.7 & 26.5 & 56.2 & 25.1 & 71.0 & -    & 49.1 \\
            Avg. & 57.2 & 22.9 & 47.1 & 17.3 & 64.7 & 46.1 & \cellcolor{Gray} \textbf{42.5} \\
            \hline
        \end{tabular}
    \end{tabular}
    }
\end{table*}

\section{Implementation details}
\label{sec:impl}

\subsection{Datasets}
\label{sub:datasets}
\noindent
We use four standard object classification benchmarks for DA to evaluate our approach. The \textbf{Office-Home} dataset \cite{office-home} consists of images from 65 categories of everyday objects from four domains - Art \textbf{(Ar)}, Clipart \textbf{(Cl)}, Product \textbf{(Pr)}, and Real World \textbf{(Rw)}. Office-31 \cite{office-31} is a simpler benchmark containing images from 31 categories belonging to three domains of objects in office settings - Amazon \textbf{(A)}, Webcam \textbf{(W)}, and DSLR \textbf{(D)}. \textbf{VisDA} \cite{visda} is a large-scale benchmark containing images from two domains - 152,397 synthetic source images and 55,388 real-world target images. Lastly, \textbf{DomainNet} \cite{M3SDA} is the largest and the most challenging dataset due to severe class imbalance and diversity of domains. It contains 345 categories of objects from six domains - Clipart \textbf{(clp)}, Infograph \textbf{(inf)}, Painting \textbf{(pnt)}, Quickdraw \textbf{(qdr)}, Real \textbf{(rel)}, Sketch \textbf{(skt)}.

\subsection{Style augmentations}
\label{sub:style}

We construct novel stylized images using 5 label-preserving augmentations on the original clean images to enable non-causal factor alignment during the training process. The augmentations are as follows:

\noindent
\begin{enumerate}[wide, labelindent=0pt]
    \item \textbf{FDA augmentation:} We use the FDA augmentation \cite{FDA} to generate stylized images based on a fixed set of style images \cite{style_imgs}. In this augmentation, a given input image is stylized by interchanging the low-level frequencies between the FFTs of the input image and the reference style image.
    \item \textbf{Weather augmentations: } We employ the frost and snow augmentations from \cite{img_aug_lib} to simulate the weather augmentation. Specifically, we use the lowest severity of frost and snow (\textit{severity} = 1) to augment the input images.
    
    \item \textbf{AdaIN augmentation: } AdaIN \cite{style_imgs} uses a reference style image to stylize a given input image by altering the feature statistics in an instance normalization (IN) layer \cite{ulyanov2017improved}. We use the same reference style image set as in FDA, and set the augmentation strength to 0.5.

    \begin{algorithm}[H]

\caption{C-SFTrans Training Algorithm}
\label{algo:overall}
\begin{algorithmic}[1]
\vspace{1mm}
\Statex \underline{\textbf{Vendor-side training}}

\vspace{1mm}
\State \textbf{Input:} Let $\mathcal{D}_s$ be the source data, $\mathcal{D}_{sty}$ be the style dataset, ImageNet pre-trained DeiT-B backbone $h$ from \cite{xu2021cdtrans}, randomly initialized goal classifier $f_g$ and randomly initialized style classifier $f_n$.

\vspace{1mm}
\Statex \underline{{\textit{Non-causal attention heads selection}}} 
\Statex \jnkc{\Comment{Fig. 3A (main paper)}}
\For{$iter < MaxTaskIters$}:
    \State Sample batch $x_i$ from $\mathcal{D}_s$
    \State Construct $x_{SCI}$ from $x_i$
    \State Compute $\mathcal{A}_i^l$ using Eq. 3 (main paper)
    \State Compute $\mathcal{L}_{cls}$ using Eq. 4 (main paper)
    \State \textbf{update} $\beta_{1j}, \beta_{2j}$ for head $j$ by minimizing $\mathcal{L}_{cls}$
\EndFor
\Statex $h_{n}^{l} = \{h : h \in h_{l}, CIS_{h} > \tau\}$

\vspace{1mm}
\For{$iter < MaxIter$}:
\vspace{1mm}
\Statex \underline{{\textit{Goal task training}}} \jnkc{\Comment{Fig. 3B (main paper)}}
\vspace{1mm}
\For{$iter < MaxTaskIters$}:
\State Sample batch from $\mathcal{D}_s$ 
\State Compute $\mathcal{L}_{cls}$ using Eq.\ 6 (main paper)
\State \textbf{update} $\theta_{h^l} \setminus \theta_{h_n^l}, ~\theta_{f_g}$ by minimizing $\mathcal{L}_{cls}$
\EndFor

\vspace{1mm}
\Statex \underline{{\textit{Style classifier training}}} \jnkc{\Comment{Fig. 3B (main paper)}}
\vspace{1mm}
\For{$iter < MaxDomainIters$}:
\State Sample batch from $\mathcal{D}_{sty}^{s}$
\State Compute $\mathcal{L}_{dom}$ using Eq.\ {1} (main paper)
\State \textbf{update} $\theta_{h_n^l}, ~\theta_{f_n}$ by minimizing $\mathcal{L}_{dom}$
\EndFor
\vspace{1mm}
\Statex \jnkc{\Comment{The two steps are carried out alternatively
}}
\EndFor

\vspace{1mm}
\Statex \underline{\textbf{Client-side training}}

\vspace{1mm}
\State \textbf{Input:} Target data $\mathcal{D}_t$, Target augmented DRI data $\mathcal{D}_t^{[i]}$, source-side pretrained backbone $h$, goal classifier $f_g$ and domain classifier $f_d$.

\vspace{1mm}
\For{$iter < MaxIter$}:
\vspace{1mm}
\Statex \underline{{\textit{Goal Task Training}}} \jnkc{\Comment{Fig. 3B (main paper)}}
\vspace{1mm}
\For{$iter < MaxTaskIters$}:
\State Sample batch from $\mathcal{D}_t$
\State Compute $\mathcal{L}_{im}$ and $\mathcal{L}_{div}$ using Eq.\ \ref{sup:eqn:loss_im}, \ref{sup:eqn:loss_div} (suppl.)
\State \textbf{update} $\theta_{h^l} \setminus \theta_{h_n^l}, ~\theta_{f_g}$ by minimizing $\mathcal{L}_{im} + \mathcal{L}_{div}$
\EndFor

\vspace{1mm}
\Statex \underline{{\textit{Style classifier training}}} \jnkc{\Comment{Fig. 3B (main paper)}}
\vspace{1mm}
\For{$iter < MaxDomainIters$}:
\State Sample batch from $\mathcal{D}_{sty}^{t}$
\State Compute $\mathcal{L}_{dom}$ using Eq.\ {1} (main paper)
\State \textbf{update} $\theta_{h_n^l}, ~\theta_{f_n}$ by minimizing $\mathcal{L}_{dom}$
\EndFor
\vspace{1mm}
\Statex \jnkc{\Comment{The two steps are carried out alternatively.
}}

\EndFor

\end{algorithmic}
\end{algorithm}
    
    \item \textbf{Cartoon augmentation: } We employ the cartoonization augmentation from \cite{img_aug_lib} to produce cartoon-style images with reduced texture from the input.
    
    \item \textbf{Style augmentation: } We use the style augmentation from \cite{jackson2019style} that augments an input image through random style transfer. This augmentation alters the texture, contrast and color of the input while preserving its geometrical features.
    
\end{enumerate}

\begin{table}[t]
    \centering
    \setlength{\tabcolsep}{7pt}
    \caption{Sensitivity analysis of alternate training on Single-Source Domain Adaptation (SSDA) on Office-Home. The goal task epochs are varied from 1 to 5.  }
    \label{tab:sensitivity_analysis}
    \vspace{-3mm}
    \resizebox{\columnwidth}{!}{%
        \begin{tabular}{cccccc}
            \toprule
            \textbf{Epochs} & \textbf{Ar $\shortrightarrow$ Cl} & \textbf{Cl $\shortrightarrow$ Pr} & \textbf{Pr $\shortrightarrow$ Rw} & \textbf{Rw $\shortrightarrow$ Ar} & \textbf{Avg.} \\
            \midrule
            1 & 63.7 & 79.8 & 79.8 & 75.7 & 74.8\\
            2 & 70.0 & 86.8 & 87.6 & 82.5 & 81.7\\
            3 & 69.9 & 86.7 & 87.5 & 82.3 & 81.6\\
            5 & 70.6 & 87.7 & 88.5 & 82.3 & 82.2\\
            \bottomrule
        \end{tabular} 
        }
    \vspace{-2mm}
\end{table}

\subsection{Experimental settings}
\label{sub:exp_set}
\vspace{-4.5pt}

In all our experiments, we use the Stochastic Gradient Descent (SGD) optimizer \cite{adam} with a momentum of 0.9 and batch size of 64. We follow \cite{SHOT} and use label smoothing in the training process. For the source-side, we train the goal task classifier for 20 epochs, and the style classifier until it achieves 80\% accuracy. On the target-side, we train the goal task classifier for 2 epochs, and use the same criteria for the style classifier as the source-side. The first 5 epochs of the source-side training are used for warm-up with a warm-up factor of 0.01. On the source-side, we use a learning rate of $8 \times 10^{-4}$ for the VisDA dataset, and $8 \times 10^{-3}$ for the remaining benchmarks. For the target-side goal task training, we use a learning rate of $5 \times 10^{-5}$ for VisDA, $2 \times 10^{-3}$ for DomainNet, and $8 \times 10^{-3}$ for the rest. Our proposed method comprises an alternate training mechanism where the goal task training and style classifier training are done alternatively for a total of 25 rounds, which is equivalent to 50 epochs of target adaptation in \cite{SHOT}. For comparisons, we implement the source-free methods DIPE \cite{wang2022exploring} and Feature Mixup \cite{kundu2022balancing} by replacing the backbone with DeiT-B. While CDTrans \cite{xu2021cdtrans} uses the entire domain for training and evaluation with the DomainNet dataset, we follow the setup of \cite{sun2022safe} to ensure fair comparisons. We train on the \textit{train} split and evaluate on the \textit{test} split of each domain.

\begin{table}[H]
    \centering
    \setlength{\tabcolsep}{10.5pt}
    \caption{Ablation study for the three components of the target-side goal task training. \textit{SSPL} denotes self-supervised pseudo-labelling.} 
    \label{tab:ablation_loss}
    \vspace{-3mm}
    \resizebox{\columnwidth}{!}{%
        \begin{tabular}{ccccl}
        \toprule
        \textbf{Method}& \textbf{$\mathcal{L}_{ent}$}& $\mathcal{L}_{div}$ & \textbf{textit{SSPL}}& \textbf{Avg.}\\
        \midrule
        Source-Only& \xmark & \xmark & \xmark & 76.4 \\
        \multirow{3}{*}{C-SFTrans} & \cellcolor{Gray} \cmark & \cellcolor{Gray} \xmark & \cellcolor{Gray} \xmark & \cellcolor{Gray}74.0 \\
                                     & \cellcolor{Gray} \cmark & \cellcolor{Gray} \cmark & \cellcolor{Gray}\xmark & \cellcolor{Gray}79.7 \color{ForestGreen}\textbf{(+5.7)} \\
                                     & \cellcolor{Gray} \cmark & \cellcolor{Gray}\cmark & \cellcolor{Gray}\cmark & \cellcolor{Gray}81.7 \color{ForestGreen}\textbf{(+7.7)} \\       
        \bottomrule
        \end{tabular}
        }
        \vspace{-3mm}
\end{table}

\section{Additional comparisons}
\label{sec:addn_comp}

We present additional comparisons with the \textbf{DomainNet} benchmark in Table \ref{tab:ssda_dnet_full}. Our method achieves the best results among the existing source-free prior arts and outperforms the source-free SHOT-B$^*$ by 3.6\%. We also observe that C-SFTrans surpasses the non-source-free method CDTrans by an impressive 5.5\%.

\begin{table}[t]
\setlength{\tabcolsep}{8pt}
\caption{Sensitivity analysis on non-causal heads (\%) for Single-Source DA on 4 settings of Office-Home}
\vspace{-2mm}
\label{tab:dom_heads}
\resizebox{\columnwidth}{!}{
\begin{tabular}{@{}cccccc@{}}
    \toprule
    $\lambda$ & \textbf{{Ar} $\shortrightarrow$ {Cl}} & \textbf{Cl $\shortrightarrow$ Pr} & \textbf{Pr $\shortrightarrow$ Rw} & \textbf{Rw $\shortrightarrow$ Ar} & \textbf{Avg.} \\ \midrule
0.1    & 70.2         & 86.7         & 87.5         & 82.4         & 81.7          \\
0.2    & 70.0         & 86.8         & 87.6         & 82.5         & 81.7          \\
0.3    & 70.3         & 86.9         & 87.7         & 82.6         & 81.9          \\
0.4    & 70.2         & 86.5         & 87.2         & 82.1         & 81.5          \\ 
\bottomrule
\end{tabular}}
\vspace{-4mm}
\end{table}

\section{Ablations on target-side goal task training}
\label{sec:abl_ta}

\noindent
\textbf{(a) Target-side goal task training loss.} Table \ref{tab:ablation_loss} shows the influence of the three loss terms in the target-side goal task training - entropy loss $\mathcal{L}_{ent}$, diversity loss $\mathcal{L}_{div}$ and self-supervised pseudo-labeling \textit{SSPL}. We observe that using $\mathcal{L}_{ent}$ alone produces lower results even compared to the source baseline. On the other hand, using both $\mathcal{L}_{ent}$ and $\mathcal{L}_{div}$ significantly improves the performance, which highlights the importance of the diversity term $\mathcal{L}_{div}$. Finally, we obtain the best results when all three components are used together for target-side adaptation, further showing the significance of the pseudo-labeling step.

\noindent
\textbf{(b) Sensitivity analysis of alternate training.} In our proposed method, we perform style classifier training and goal task training in an alternate fashion, \ie the task classifier $f_g$ is trained for a few epochs, followed by the training of the style classifier $f_n$ until it reaches a certain accuracy threshold (empirically set to $80\%$). In Table \ref{tab:sensitivity_analysis}, we show the effect of varying the number of epochs of the goal task training from 1 to 5, and observe the impact on the goal task accuracy during non-causal factor alignment. We observe that 2 epochs of goal task training achieves the optimal balance between target accuracy and training effort.  We observe that just a single epoch of task classifier training negatively impacts the goal task performance. While 3 epochs achieves the best performance, it involves significant training effort for merely $0.5\%$ improvement in the task accuracy. Therefore, 2 epochs of goal task training achieves the optimal balance between target accuracy and training effort. 

\noindent
\textbf{(c) Selection of non-causal heads.}
We select a set of non-causal attention heads based on their \textit{Causal Influence Score} (CIS). We sort the CIS in descending order and select the top $\lambda$\% of heads satisfying the condition $\textit{\text{CIS}} > \tau$. In Table \ref{tab:dom_heads}, we present the effect of altering this hyperparameter $\lambda$ on the overall performance. We observed that with a lower value of $\lambda$, the pathways formed by non-causal heads do not adequately extract and learn the non-causal factors, which consequently hinders the domain-invariant alignment and leads to non-optimal task performance. Similarly, increasing $\lambda$ too much reduces the ability of the network to learn causal factors and leads to lower performance. Overall, our approach is not very sensitive towards this hyperparameter.

\begin{table}[h]
\vspace{-2mm}
\setlength{\tabcolsep}{8pt}
\caption{Ablation study for the effect of augmentations for target-side goal task training. } 
\label{tab:augs}
\resizebox{\columnwidth}{!}{
\begin{tabular}{@{}cccccc@{}}
    \toprule
    \textbf{No. of augs.} & \textbf{Ar $\shortrightarrow$ Cl} & \textbf{Cl $\shortrightarrow$ Pr} & \textbf{Pr $\shortrightarrow$ Rw} & \textbf{Rw $\shortrightarrow$ Ar} & \textbf{Avg.} \\ \midrule
    3 & 64.3 & 79.9 & 84.6 & 80.0 & 77.2 \\
    6 & 70.0 & 86.8 & 87.6 & 82.5 & 81.7 \\ \bottomrule
\end{tabular}}
\vspace{-3mm}
\end{table}

\noindent
\textbf{(d) Effect of augmentations.} Table \ref{tab:augs} demonstrates that fewer augmentations for the style classifier significantly deteriorate the adaptation performance in comparison to the full set of augmentations. This indicates that a more complex style classification task better facilitates the non-causal factor alignment. However, due to the scarcity of more complex augmentations, we use the six outlined in Sec. \ref{sub:style}

{\small
\bibliographystyle{ieee_fullname_fix}
\bibliography{main}
}

\end{document}